\documentclass[conference]{IEEEtran}
\IEEEoverridecommandlockouts
\usepackage{amsmath,amssymb,amsfonts}
\usepackage{algorithmic}
\usepackage{graphicx}
\usepackage{textcomp}
\usepackage{xcolor}
\usepackage{tabularx}
\usepackage{ragged2e}
\usepackage{array}
\usepackage[table]{xcolor}
\usepackage{colortbl}
\usepackage{url}
\usepackage{tikz}
\usetikzlibrary{arrows.meta,positioning,calc}
\usepackage[most]{tcolorbox}
\tcbuselibrary{listings,breakable}
\usepackage{xcolor}

\definecolor{bestgreen}{RGB}{200,230,201}
\definecolor{secondorange}{RGB}{255,224,178}

\def\BibTeX{{\rm B\kern-.05em{\sc i\kern-.025em b}\kern-.08em
    T\kern-.1667em\lower.7ex\hbox{E}\kern-.125emX}}
\begin{document}

\title{{Systematic Evaluation of Large Language Models for Post-Discharge Clinical Action Extraction}

}

\author{
\IEEEauthorblockN{Shivali Dalmia\IEEEauthorrefmark{1}}
\IEEEauthorblockA{\textit{Centific AI Research} \\
\textit{Centific Global Solutions, Inc.}\\
shivali.dalmia@centific.com}
\and
\IEEEauthorblockN{Ananya Mantravadi\IEEEauthorrefmark{1}}
\IEEEauthorblockA{\textit{Centific AI Research} \\
\textit{Centific Global Solutions, Inc.}\\
ananya.mantravadi@centific.com}
\and
\IEEEauthorblockN{Prasanna Desikan}
\IEEEauthorblockA{\textit{Centific AI Research} \\
\textit{Centific Global Solutions, Inc.}\\
prasanna.desikan@centific.com}
\thanks{\IEEEauthorrefmark{1}These authors contributed equally to this work. }
}
% The order of the authors’ names in this paper has been decided by tossing a coin.

\maketitle

\begin{abstract}
The work in this paper evaluates zero-shot and few-shot large language models (LLMs) for safety-critical clinical action extraction using the CLIP discharge-note dataset, with particular emphasis on transitions of care and post-discharge patient safety. To manage the complexity of clinical documentation, we introduce a two-stage extraction framework that decomposes discharge notes, that are written in narrative form,  into fine-grained, explicitly actionable clinical tasks through a staged prompting strategy. Our contributions include a systematic assessment of generative LLMs for clinical action extraction, a detailed comparison between general-purpose LLMs and task-specific supervised BERT-based models, and an analysis of annotation inconsistencies across different action categories.

We show that contemporary LLMs achieve performance comparable to or exceeding supervised models on binary actionability detection, while supervised baselines retain a meaningful advantage on fine-grained multi-label category classification, despite the absence of task-specific fine-tuning and under strict data-privacy constraints.  Qualitative error analysis reveals that many failures stem from misalignment between model reasoning and dataset annotation conventions, particularly in cases involving implicit clinical actions and rigid structural labeling rules. These results indicate that reported performance reflects  model limitations due to lack of clinical reasoning, that is not captured by plain annotations. Labels without rationales make it impossible to distinguish clinical reasoning failures from annotation convention mismatches. Advancing clinical NLP requires reasoning-annotated datasets that document \textit{why} specific spans are actionable, not merely \textit{which} spans were labeled, enabling proper evaluation of model clinical understanding. Integrating LLM-based extraction with specialized classifiers and human-in-the-loop validation enable more reliable, deployment-ready medical agent workflows.
\end{abstract}

\begin{IEEEkeywords}
Large language models, Clinical information extraction, Discharge summaries, Care transitions, Natural language processing, Medical informatics
\end{IEEEkeywords}

\section{Introduction}

The patient journey can be defined as the sequence of a patient's interactions with the healthcare system over time, reflecting how care progresses across multiple clinical settings as health needs change. Rather than a single episode, this journey unfolds longitudinally, moving from primary care and ambulatory clinics to acute care environments, such as the Emergency Room (ER), during periods of urgent clinical deterioration. Each setting operates under distinct goals, workflows, and documentation practices, making the transfer of information between them a critical vulnerability. Among these transitions, the shift from hospital to home has been consistently identified as a period of elevated risk, where the safety and effectiveness of care depend heavily on the clarity of discharge documentation.

Discharge summaries are intended to bridge this gap by summarizing the hospital course and outlining the next steps for post-discharge care. However, effective care transitions face a dual challenge. First, from the patient perspective, Coleman et al.~\cite{coleman2013understanding} found that many older patients return home with limited understanding of their discharge instructions, often unable to recall their discharge diagnoses, treatment plans, or medication regimens. Their work identified health literacy, cognition, and self-efficacy as main predictors of instruction comprehension, underscoring that the challenge extends beyond simply identifying actionable items to ensuring they are communicated effectively. Second, from the clinician perspective, prior studies~\cite{mullenbach-etal-2021-clip,desai2021empowering} have shown that discharge notes are often lengthy, complex, and inconsistently structured, containing substantial copied or historical content that obscures information most relevant for future care. This lack of structure imposes significant cognitive burden on downstream Primary Care Providers (PCPs), who must manually sift through narrative text to identify actionable items such as pending test results, medication adjustments, and conditional follow-up plans. These patient comprehension failures and clinician extraction challenges lead to adverse drug events, delayed care, and preventable readmissions, with annual Medicare costs estimated at \$17 billion \cite{coleman2013understanding}.

\subsection{Motivation}

Recent advances in Artificial Intelligence (AI) have motivated increasing interest in automated analysis of discharge documentation to address these extraction challenges. Early approaches focused on supervised, task-specific extraction using encoder-only models trained on curated clinical datasets, such as the CLIP dataset~\cite{mullenbach-etal-2021-clip} for follow-up action identification across seven clinical action categories: Appointment, Medication, Lab, Imaging, Procedure, Patient Instructions, and Other (see table \ref{tab:action_categories}). Subsequent efforts expanded task coverage to include clinical decision rationale extraction~\cite{elgaar-etal-2024-meddec}. More recently, Large Language Models (LLMs) have been explored for discharge summarization and hospital course documentation~\cite{zhou2024surveylargelanguagemodels}. While these models offer promise in reducing documentation burden without the need for task-specific training, they introduce new safety risks, like hallucination and unreliable extraction of follow-up actions.

This reinforces the need for rigorous evaluation of both specialized and general-purpose models on the safety-critical task of extracting post-discharge follow-up actions from real-world clinical documentation. No prior work has systematically compared modern LLMs against established supervised baselines specifically for this task, to the best of our knowledge. This paper addresses that gap by benchmarking five general-purpose generative models---including GPT-5.2, Gemini-3-Flash, Claude Sonnet 3.5, DeepSeek-V3.2, and medical-specialized MedGemma-27B-it---against supervised BERT baselines on the CLIP dataset. We aim to determine whether modern LLMs can extract post-discharge interventions effectively and safely, offering a flexible alternative to traditional supervised pipelines while strictly adhering to privacy standards required for responsible processing of sensitive healthcare data.

\subsection{Contributions}

% The main contributions of this work are as follows:
\begin{itemize}
    \item We propose a two-stage extraction framework that decomposes clinical documentation into fine-grained, directly actionable tasks.
    \item We conduct a systematic evaluation of zero-shot and few-shot generative language models on clinical action extraction.
    \item We compare general-purpose pretrained language models against specialized supervised BERT architectures on safety-critical extraction tasks.
    \item We provide category-level performance analysis and identify systematic annotation inconsistencies across discharge-related task types.
\end{itemize}

% \section{Motivation}

% Transitions from hospital to outpatient care are high-risk periods in the patient journey, where missed or unclear follow-up actions can lead to medication errors, delayed care, and preventable readmissions. Although discharge notes are intended to coordinate these transitions, they are often long, inconsistently structured, and difficult to interpret under routine clinical time constraints. As a result, clinically important next steps may be overlooked, particularly when responsibilities span multiple providers and settings.

% Automated extraction of post-discharge actions has the potential to reduce this burden and improve continuity of care. While prior work has demonstrated the feasibility of extracting follow-up tasks from discharge documentation, most evaluations have focused on specialized, supervised models. With the increasing availability of large language models, there is a need for systematic evidence on whether general-purpose generative models can support this safety-critical workflow without introducing new risks.

\section{Problem Formulation}

We formulate post-discharge clinical action extraction as a benchmarking problem over unstructured discharge notes. Given a discharge document, the task is to identify spans corresponding to fine-grained, patient-specific action items that are directly executable after discharge, such as follow-up appointments, laboratory tests, medication-related actions, and conditional monitoring instructions. Action items are distinguished from contextual or historical statements by whether they require action to be taken after discharge.

We use the CLIP dataset, which provides expert-annotated discharge notes with labeled action spans across predefined task categories~\cite{mullenbach-etal-2021-clip}. Within this framework, we evaluate and compare task-specific encoder-based models and generative transformer-based language models operating in zero-shot and few-shot settings. Evaluation focuses on extraction accuracy, category-level performance, and failure modes relevant to safety-critical clinical workflows.
\begin{table}[t]
\centering
\caption{Clinical Action Categories and Definitions}
\label{tab:action_categories}
\footnotesize
\begin{tabular}{p{3.4cm} p{4.8cm}}
\hline
\textbf{Action Category} & \textbf{Definition} \\
\hline
Appointment-related follow-up 
& Instructions to schedule or attend a clinical visit, consultation, or referral. \\
\hline
Lab-related follow-up 
& Instructions to order, repeat, or review laboratory tests. \\
\hline
Medication-related follow-up 
& Instructions to start, stop, change, hold, or adjust medications. \\
\hline
Imaging-related follow-up 
& Instructions to schedule or obtain diagnostic imaging studies. \\
\hline
Procedure-related follow-up 
& Instructions related to medical or surgical procedures and their care. \\
\hline
Case-specific instructions for patient 
& Direct care instructions given to the patient. \\
\hline
Other helpful contextual information 
& Clinically useful information that supports follow-up or care coordination. \\
\hline
\end{tabular}
\end{table}

\section{Related Work}

The landscape of clinical natural language processing has undergone a fundamental shift in recent years. Early approaches relied heavily on encoder-only architectures such as BioBERT~\cite{biobert}, ClinicalBERT~\cite{clinicalbert}, and domain-adapted BERT variants trained on PubMed \cite{canese2013pubmed} abstracts or MIMIC clinical notes \cite{PhysioNet-mimiciii-1.4_database}. These models achieved strong performance on clinical NLP tasks but required task-specific fine-tuning for each new application. As documented in recent surveys~\cite{zhou2024surveylargelanguagemodels}, the emergence of large-scale decoder-only models such as GPT-4, Med-PaLM 2, and their successors has enabled a paradigm shift from fine-tuning to prompting-based approaches. While BERT-based models excel at high-volume, well-defined tasks with abundant training data, modern LLMs demonstrate superior performance on unstructured and ambiguous clinical narratives where contextual reasoning is required. 
% However, systematic evaluation of these models on safety-critical clinical extraction tasks remains limited.

Clinical information extraction has traditionally focused on identifying structured entities (medications, diagnoses, procedures) and their relationships within clinical text. Early work used rule-based systems and conditional random fields, and large-scale annotated corpora have been developed to support these efforts~\cite{patel2018annotation,nye2018corpus}. 

Recent efforts have expanded beyond entity recognition to higher-level semantic tasks. Frameworks such as COMCARE~\cite{jin2025comcare} have demonstrated that ensemble approaches combining context-aware named entity recognition with relation extraction can achieve high accuracy on structured EHR data. MedDecXtract~\cite{elgaar-etal-2025-meddecxtract} introduced a clinician-facing system for extracting and visualizing medical decision statements from discharge summaries  according to DICTUM (Decision Identification and Classification Taxonomy for Use in Medicine), covering ten decision categories including diagnoses, treatment goals, and test orders. It focuses on understanding clinical decision-making patterns across the entire hospital course.
% , and does not specifically target post-discharge actionable follow-up items, which require distinguishing future-oriented instructions from historical clinical reasoning.

Discharge summaries represent a particularly challenging document type due to their length, narrative structure, and mixing of historical context with future-oriented instructions. The CLIP dataset~\cite{mullenbach-etal-2021-clip} pioneered the task of extracting follow-up action items from discharge notes, providing expert annotations across seven clinical action categories: Appointment, Medication, Lab, Imaging, Procedure, Patient Instructions, and Other. The original work established supervised BERT-based baselines achieving Macro F1 scores of 0.661--0.668 through domain adaptation and contextual encoding. These models require substantial annotated training data and task-specific fine-tuning, limiting their adaptability to evolving clinical workflows or new healthcare settings. 

Despite the proliferation of medical LLM benchmarks evaluating question-answering~\cite{zuo2025medxpertqa}, diagnostic reasoning~\cite{wang2025medagent}, and multi-step clinical workflows~\cite{jiang2025medagentbench,mantravadi2026art}, a critical gap remains: no prior work has rigorously benchmarked general-purpose LLMs against supervised baselines specifically for post-discharge actionable follow-up items extraction. 

% Complementary work has explored discharge summarization to condense lengthy narratives for physician review~\cite{adams2021summary,pivovarov2015summarization}, often using extractive methods~\cite{alsentzer2018extractive} or problem-oriented approaches~\cite{gao2022summarizing,liang2021problem}. Interactive visualization systems like HARVEST~\cite{hirsch2014harvest} have demonstrated the value of longitudinal patient record summarization with temporal organization. More recent efforts have focused on extracting clinical decision rationales~\cite{elgaar-etal-2024-meddec} and patient-centered treatment goals~\cite{bojic-etal-2025-smartminer}.

Our work addresses this gap through: (1) systematic comparison of five modern LLMs, including general-purpose and medical-specialized variants, against established BERT baselines on the CLIP dataset, (2) decomposition of the task into explicit actionability filtering and category classification stages to isolate performance patterns, (3) strict adherence to Zero Data Retention protocols for privacy-compliant evaluation, and (4) detailed error analysis revealing that apparent model failures often reflect annotation inconsistencies rather than understanding deficits. This evaluation provides the first comprehensive assessment of whether prompt-driven LLM approaches can serve as flexible, deployment-efficient alternatives to traditional supervised pipelines for safety-critical clinical extraction tasks.

\section{Approach}

% ---------- High-level workflow diagram (Approach + Execution) ----------
\begin{figure*}[h!]
    \centering % Centers the image within the figure environment
    \includegraphics[width=1\textwidth]{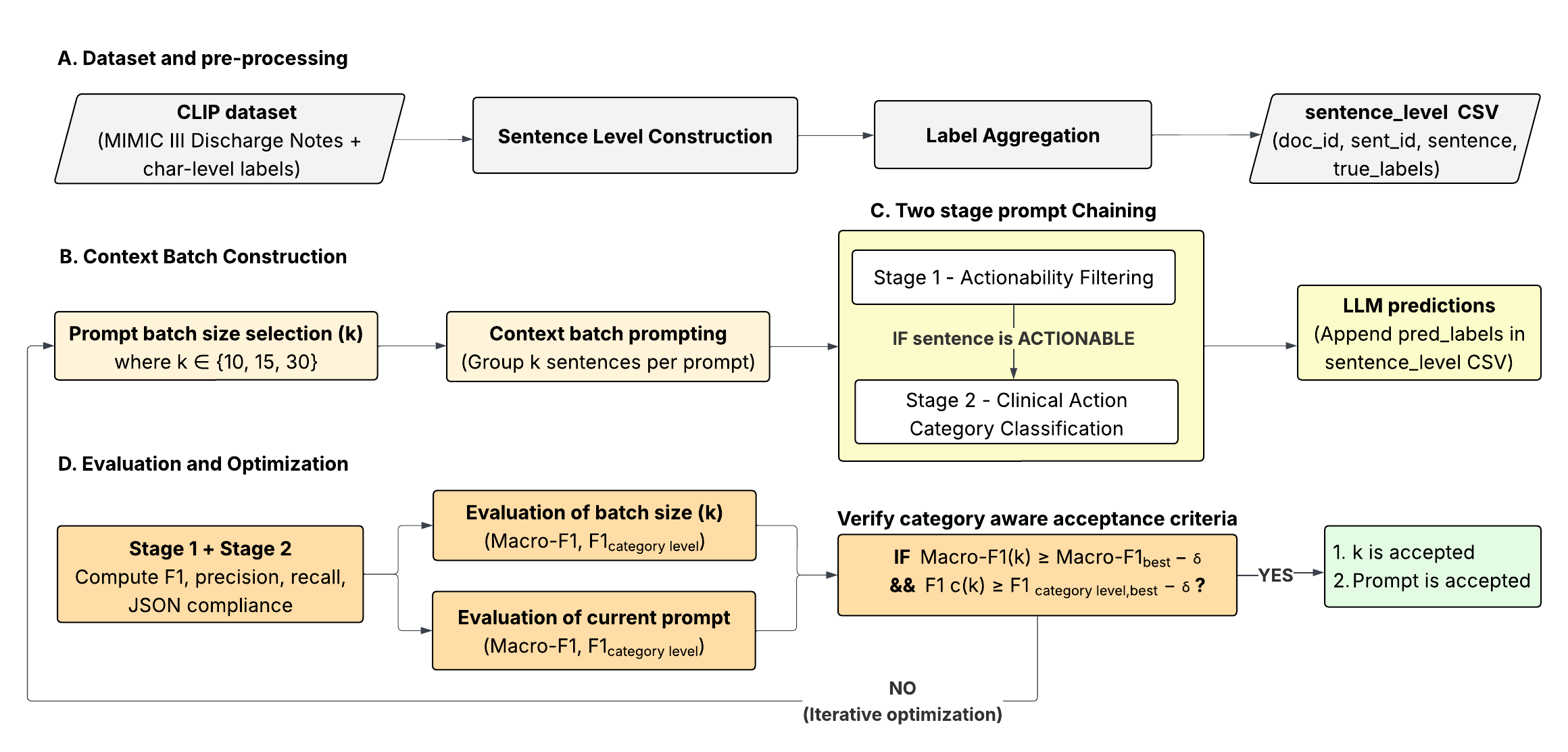} % Include your image
    \caption{High-level workflow of the proposed LLM-driven annotation framework. Discharge notes are converted to sentence-level instances, processed in context-batched prompts of size \(k\), and labeled using two-stage prompt chaining. Performance is quantified using precision, recall, F1 and JSON compliance, and used to optimize \(k\) and refine prompts iteratively} % Caption for the figure
    \label{fig:approach} % Label for cross-referencing
% \vspace{-10pt}
\end{figure*}

We propose an end-to-end LLM-driven annotation framework for identifying and categorizing actionable clinical follow-up tasks for primary care physicians from discharge summary notes. As described in Figure \ref{fig:approach}, our approach reformulates the problem as a sentence-level classification task and combines structured data preprocessing, context-aware prompting, and prompt chaining to improve robustness and clinical relevance. To reduce spurious predictions from descriptive clinical narratives, we adopt a two-stage prompting strategy that explicitly separates actionability detection from task type classification. The framework is further refined through iterative prompt optimization guided by quantitative evaluation and systematic error analysis.

\subsection{Dataset and pre-processing}

\subsubsection{\textbf{Dataset}}  We use the CLIP dataset~\cite{mullenbach-etal-2021-clip}, which provides expert-annotated discharge summaries from the MIMIC-III critical care database~\cite{PhysioNet-mimiciii-1.4_database, PhysioNet-mimiciii-1.4_paper}, a large, de-identified dataset hosted on PhysioNet.
The CLIP annotations are provided at the character-span and sentence-span levels, where each span is associated with one or more clinical task labels.
\subsubsection{\textbf{Sentence-level construction}}
Since actionable clinical follow-up tasks are naturally expressed at the sentence level, we convert the original character-level annotations into a sentence-level representation following the official pre-processing pipeline provided in the CLIP repository. Each discharge summary is first segmented into sentences. For each sentence, all character-level annotation spans that overlap with its character offsets are identified. An annotated span is the exact piece of text that a human marks as expressing a specific clinical action, with defined start and end boundaries, serving as the most precise ground-truth signal in datasets like CLIP. For example, in the discharge note sentence \textit{“The patient was advised to avoid peanuts and use the albuterol inhaler every 4 hours as needed for wheezing”}, the two annotated spans are \textit{avoid peanuts} and \textit{use the albuterol inhaler every 4 hours as needed for wheezing}, which are selected because they contain explicit, executable instructions. These spans are then assigned labels reflecting the type of action, such as case-specific patient instruction and medication-related follow-up, while the rest of the sentence remains unannotated.

\subsubsection{\textbf{Label aggregation}} Label aggregation explains how detailed span annotations are converted into sentence-level ground truth labels. For each sentence in a discharge note, all annotated spans whose character boundaries intersect with that sentence are identified. Using the allergy and asthma example, the sentence \textit{“The patient was advised to avoid peanuts and use the albuterol inhaler every 4 hours as needed for wheezing”} contains two annotated spans within the same sentence. Because both spans appear in this sentence, their labels are collected into a single list of true labels for that sentence. Since the two spans represent different categories of actions, one being a \textit{case specific instruction for the patient}  and the other a \textit{medication-related follow up}, the sentence is treated as a multi label instance. By contrast, a sentence that contains no annotated spans is assigned an empty label list, indicating that it does not contain any annotated clinical follow up actions. Actionable content is sparse in discharge summaries: only 1,771 of 15,591 test sentences (11.36\%) contain labeled action items.

\subsubsection{\textbf{Output format}} The processed data are exported into a sentence-level CSV file, where each row contains:
\begin{itemize}
    \item \textit{doc\_id}: an identifier linking the sentence to its source note (e.g., 45901).
    \item \textit{sent\_index}: the index of the sentence within the document (e.g., 143).
    \item \textit{sentence}: the tokenized sentence text (e.g., ['Discharge','Disposition','Home','With','Service']).
    \item \textit{true\_labels}: a list of ground-truth labels associated with the sentence (e.g.,['Appointment-related followup']).
\end{itemize}

This transformation preserves the original CLIP annotations while standardizing the unit of analysis from character spans to sentences. The resulting sentence-level dataset enables simplified evaluation, and directly support sentence-based actionability filtering and category classification. 

\subsection{Context Batch Construction}

Discharge summaries often express actionable follow-up instructions using contextual dependencies that span multiple consecutive sentences, such as section headers (e.g., \textit{Discharge Instructions}, \textit{Follow-up Instructions}) and references to previously mentioned medications, tests, or timelines. Interpreting sentences in isolation can therefore lead to missed or misclassified clinical actions.

\subsubsection{\textbf{Context Batched Prompting}}

We adopt a context-batched prompting strategy in which a fixed number of contiguous sentences are grouped and jointly provided as input to the LLM. Let
\[
S = \{ s_1, s_2, \dots, s_n \}
\]
denote the ordered sequence of sentences extracted from a discharge summary note. The sentence sequence is partitioned into batches of size \(k\), where each batch is defined as
\[
W_j = \{ s_{(j-1)k+1}, s_{(j-1)k+2}, \dots, s_{jk} \}
\]
for \(j = 1, 2, \dots, \lfloor n/k \rfloor\), with the final batch containing the remaining sentences if \(n\) is not divisible by \(k\). We do not overlap sentences between batches.

% where each batch is defined as
% \[
% W_j = \{ s_{(j-1)k+1}, s_{(j-1)k+2}, \dots, s_{jk} \}.
% \]
Here, \(k\) denotes the prompt batch size, i.e., the number of sentences included in a single prompt. In our experiments,
\[
k \in \{10, 15, 30\}.
\]

For each batch \(W_j\), the model is conditioned on all \(k\) sentences and generates predictions independently for every sentence within the batch. This design allows the model to use information from nearby sentences as context while still making predictions separately for each sentence.

\subsubsection{\textbf{Prompt Batch Size Selection}}

The prompt batch size \(k\) is treated as a tunable hyperparameter and optimized using a \textbf{successive halving strategy}. Candidate values are selected from the discrete set $k \in \{10, 15, 30\}$.

At every stage, decisions to retain or discard a value of \(k\) are based on three parameter selection criteria evaluated for each model:
(i) \textbf{Macro-F1} \textit{(the average F1 score computed across all seven clinical action categories, treating each category equally)},  
(ii) \textbf{Per-category F1 scores} \textit{(the individual F1 score for each clinical action category)}, and  
(iii) \textbf{Strict JSON compliance} \textit{(the percentage of outputs that follow the required structured format)}.  

JSON compliance is treated as a hard validity constraint: any configuration that fails to produce valid structured outputs is discarded. Macro-F1 and per-category F1 scores are evaluated using the category-aware acceptance criteria defined in the following section.

\begin{itemize}
    \item \textbf{Stage 1 – Preliminary Screening:}  
    All values of \(k\) are evaluated on a small subset of the training data (e.g., 50 discharge summary documents). For each value of \(k\), macro-F1, per-category F1 scores, and JSON compliance are computed. A value of \(k\) is discarded if it fails JSON compliance or if its macro-F1 or per-category F1 scores violate the acceptance thresholds defined by the category-aware criteria.

    \item \textbf{Stage 2 – Refined Evaluation:}  
    The remaining values of \(k\) are evaluated on a larger subset of the training data (e.g., 100 discharge summary documents). The same three criteria — macro-F1, per-category F1, and JSON compliance are re-evaluated, and configurations that no longer satisfy the acceptance criteria are further eliminated.

    \item \textbf{Stage 3 – Final Selection:}  
    The retained value(s) of \(k\) are evaluated on the full training set (e.g., 518 discharge summary documents). The final prompt batch size is selected to maximize macro-F1 while satisfying all per-category F1 constraints and maintaining consistent JSON formatting compliance.
\end{itemize}

This successive halving procedure ensures that prompt batch size selection is driven by balanced performance across categories and stable, machine-consumable outputs, rather than aggregate accuracy alone.

\begin{table*}[t]
\centering
\caption{Examples for Stage 1 (Actionability Filtering) and Stage 2 (Action Category Classification)}
\label{tab:actionable_examples}
\begin{tabular}{p{6cm} p{2.5cm} p{4cm} p{4cm}}
\hline
\textbf{Sentence} & \textbf{Stage 1} & \textbf{Stage 2} & \textbf{Reason} \\
\hline
\textit{The patient was instructed to hold ASA and refrain from NSAIDs for 2 weeks.}
& Actionable 
& Medication-related follow-up
& Explicit medication change/hold instruction \\
\hline

\textit{Discharged on Atorvastatin 40 mg daily and Aspirin 81 mg.}
& Non-Actionable 
& -- 
& Standard continuation of existing medications \\
\hline

\textit{The patient requires a neurology consult at XYZ for evaluation.}
& Actionable 
& Appointment-related follow-up
& Explicit specialist referral \\
\hline

\textit{Follow up with your primary care provider as needed for any new concerns.}
& Non-Actionable
& -- 
& Non-specific, conditional guidance without a concrete action \\
\hline

\textit{We ask that the patient's family physician repeat these tests in 2 weeks to ensure resolution.}
& Actionable
& Lab-related follow-up
& Explicit laboratory test follow-up \\
\hline

\textit{Please arrive at 11 am for X-rays before your visit.}
& Actionable
& Appointment-related and imaging-related follow-up
& Multiple explicit procedural actions \\
\hline
\end{tabular}
\end{table*}

\subsubsection{\textbf{Category-Aware Acceptance Criterion}}

A category-aware acceptance criterion is used to balance overall performance with stability across the seven clinical action categories. Overall performance is measured using the \textbf{Macro-F1} score, together with a “must-not-degrade” constraint that limits performance drops at the individual category level. These criteria are applied at every stage of the successive halving procedure described above.

\begin{itemize}
    \item \textbf{Macro-F1 tolerance:}  
    A parameter setting is considered acceptable only if its macro-F1 score is within a fixed tolerance of the best-performing setting at the current stage. In our experiments, this tolerance is set to 5--10\% relative difference:
    \[
    \text{Macro-F1}(k) \ge \text{Macro-F1}_{\text{best}} - \delta .
    \]
    where \(\delta \in [0.05-0.10]\)
    \item \textbf{Per-category degradation tolerance:}  
    For each of the seven clinical action categories \(c\), the F1 score must not decrease by more than a small margin \(\delta\) relative to the best-performing configuration for that category:
    \[
    F1_c(k) \ge F1_{c,\text{best}} - \delta ,
    \]
    where \(\delta \in [0.05-0.10]\).
\end{itemize}

Formally, a prompt batch size \(k\) is accepted only if:
\[
\text{Macro-F1}(k) \ge \text{Macro-F1}_{\text{best}} - \delta
\]
and for every category \(c\),
\[
F1_c(k) \ge F1_{c,\text{best}} - \delta .
\]

This criterion prevents a configuration from being selected based solely on strong performance in frequent or less critical categories (e.g., routine appointments) while degrading performance in clinically sensitive categories such as medication changes, laboratory tests, imaging follow-ups, or referrals. It ensures that improvements in overall performance reflect consistent and reliable behavior across all clinical action categories.

\subsection{\textbf{Two-Stage Prompt Chaining}}
We formulate actionable task extraction as a two-step reasoning problem and implement it using prompt chaining. Instead of directly classifying every sentence into one of the clinical action categories, we decompose the task into (i) identifying whether a sentence is actionable and (ii) assigning an appropriate action category only to those sentences that are deemed actionable. This separation improves robustness by reducing false positives from descriptive or narrative clinical text.
\subsubsection{\textbf{Stage 1: Actionability Filtering}}
In the first stage, each sentence is evaluated to determine whether it expresses a concrete clinical follow-up action. A sentence is labeled \textbf{\textit{Actionable}} if it contains an explicit or implicit instruction requiring an action such as a medication change, laboratory test, referral, or procedural step. For example, instructions to hold medications (e.g., \textit{``hold ASA and refrain from NSAIDs''}), request laboratory follow-ups (e.g., \textit{``repeat these tests in 2 weeks''}), or specify procedural actions (e.g., \textit{``arrive at 11 am for X-rays''}) are treated as actionable, as shown in Table~\ref{tab:actionable_examples}.

A sentence is labeled \textbf{\textit{Non-Actionable}} if it is descriptive, maintains the status quo, or lacks a concrete instruction. This includes medication continuation without change (e.g., \textit{``Discharged on Atorvastatin 40 mg daily and Aspirin 81 mg''}) and non-specific guidance (e.g., \textit{``follow up as needed''}). 

This filtering step removes descriptive and status-quo content before category classification, reducing false positives and improving the reliability of downstream action item categorization.

% -------- Iterative prompt reginement diagram-----------%
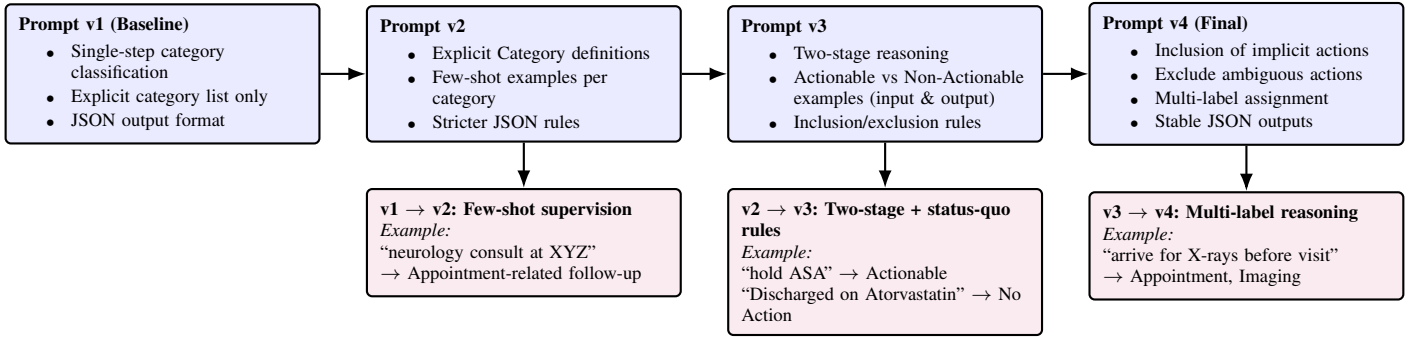
\begin{figure*}[t]
\centering
\begin{tikzpicture}[
  font=\scriptsize,
  node distance=6mm and 6mm,
  box/.style={
    rounded corners=2pt,
    draw,
    thick,
    align=left,
    inner xsep=5pt,
    inner ysep=5pt,
    text width=#1
  },
  vbox/.style={box=3.8cm, fill=blue!8},
  rbox/.style={box=3.8cm, fill=purple!8},
  arrow/.style={-Latex, thick}
]

% -------------------------
% Version boxes (top row)
% -------------------------
\node[vbox] (v1) {\textbf{Prompt v1 (Baseline)}\\
\begin{itemize}\setlength\itemsep{1pt}
\item Single-step category classification
\item Explicit category list only
\item JSON output format
\end{itemize}};

\node[vbox, right=of v1] (v2) {\textbf{Prompt v2}\\
\begin{itemize}\setlength\itemsep{1pt}
\item Explicit Category definitions
\item Few-shot examples per category
\item Stricter JSON rules
\end{itemize}};

\node[vbox, right=of v2] (v3) {\textbf{Prompt v3}\\
\begin{itemize}\setlength\itemsep{1pt}
\item Two-stage reasoning
\item Actionable vs Non-Actionable examples (input \& output)
\item Inclusion/exclusion rules
\end{itemize}};

\node[vbox, right=of v3] (v4) {\textbf{Prompt v4 (Final)}\\
\begin{itemize}\setlength\itemsep{1pt}
\item Inclusion of implicit actions
\item Exclude ambiguous actions
\item Multi-label assignment
\item Stable JSON outputs
\end{itemize}};

% -------------------------
% Refinement boxes (bottom row)
% -------------------------
\node[rbox, below=of v2] (r12) {\textbf{v1 \(\rightarrow\) v2: Few-shot supervision}\\
\textit{Example:}\\
``neurology consult at XYZ''\\
\(\rightarrow\) Appointment-related follow-up};

\node[rbox, below=of v3] (r23) {\textbf{v2 \(\rightarrow\) v3: Two-stage + status-quo rules}\\
\textit{Example:}\\
``hold ASA'' \(\rightarrow\) Actionable\\
``Discharged on Atorvastatin'' \(\rightarrow\) No Action};

\node[rbox, below=of v4] (r34) {\textbf{v3 \(\rightarrow\) v4: Multi-label reasoning}\\
\textit{Example:}\\
``arrive for X-rays before visit''\\
\(\rightarrow\) Appointment, Imaging};

% -------------------------
% Arrows between versions
% -------------------------
\draw[arrow] (v1.east) -- (v2.west);
\draw[arrow] (v2.east) -- (v3.west);
\draw[arrow] (v3.east) -- (v4.west);

% -------------------------
% Arrows from versions to refinements
% -------------------------
\draw[arrow] (v2.south) -- (r12.north);
\draw[arrow] (v3.south) -- (r23.north);
\draw[arrow] (v4.south) -- (r34.north);

\end{tikzpicture}

\caption{Iterative prompt refinement: Each iteration introduces targeted changes by error driven analysis. One representative example is shown per refinement step, illustrating how added supervision, task decomposition, and rule-based constraints improve action detection and category classification.}
\label{fig:iterative_prompt_refinement}
\end{figure*}
%---------------------------------------------
\subsubsection{\textbf{Stage 2: Clinical Action Category Classification}}
In Stage 2, only sentences that are labeled as Actionable in Stage 1 are passed for classification into one or more predefined clinical action categories as described in Table ~\ref{tab:action_categories}. This stage focuses on identifying the specific type of follow-up action expressed in the sentence. The prompt is designed to include:
\begin{itemize}
    \item clear and precise definitions of the seven clinical action categories,
    \item representative positive and negative examples,
    \item explicit inclusion and exclusion criteria, and
    \item strict instructions for producing structured JSON-formatted outputs.
\end{itemize}

For example, medication-related instructions such as \textit{``The patient was instructed to hold ASA and refrain from NSAIDs for 2 weeks''} are classified as \textit{Medication-related follow-up}, while sentences involving multiple actions, such as \textit{``Please arrive at 11 am for X-rays before your visit''}, are assigned multiple categories (e.g., \textit{Appointment-related} and \textit{Imaging-related follow-up}). Representative examples of these mappings are shown in Table~\ref{tab:actionable_examples}. Sentences labeled as \textit{Non-Actionable} in Stage~1 are excluded from Stage~2 and receive no category labels. For instance, \textit{``Discharged on Atorvastatin 40 mg daily and Aspirin 81 mg''} describes medication continuation without any required follow-up action and is therefore filtered out, as illustrated in Table~\ref{tab:actionable_examples}.

This design ensures that Stage~2 operates only on clinically meaningful instructions and naturally supports multi-label classification when a sentence contains multiple distinct follow-up actions.

\subsubsection{\textbf{Iterative Prompt Refinement}}

We refine the prompts iteratively using both quantitative evaluation and qualitative error analysis. After each prompt version, we analyze false positives, false negatives, and category-level confusion patterns across both stages, and use these findings to update category definitions, add clarifying examples, and strengthen inclusion and exclusion rules. This process is repeated until performance stabilizes and further improvements become marginal. As described in Figure~\ref{fig:iterative_prompt_refinement}, the final prompts therefore reflect multiple refinement cycles guided by empirical evaluation.

\subsection{Evaluation Protocol}

The proposed approach is evaluated to quantify its effectiveness in extracting actionable clinical follow-up tasks and assigning accurate action categories at the sentence level. The protocol defines how results are generated, reported, and compared across models, prompt batch sizes, and prompt versions.

\subsubsection{\textbf{Stage-wise Performance Reporting}}

Performance is reported separately for each stage of the pipeline.  

For \textbf{Stage 1 (Actionability Filtering)}, precision, recall, and F1 are reported for the \textit{Actionable} and \textit{Non-Actionable} classes, together with Macro-F1, measuring how well models distinguish clinical actions from descriptive content.  

For \textbf{Stage 2 (Clinical Action Category Classification)}, precision, recall, and F1 are reported for each of the seven action categories, along with Macro-F1 to summarize balanced category-level performance. All metrics are computed at the sentence level.

\subsubsection{\textbf{Evaluation of Prompt Batch Size}}

For each model, the full two-stage pipeline is executed independently for every prompt batch size \(k \in \{10, 15, 30\}\). For each configuration, Stage-wise F1 scores, Macro-F1, per-category F1, and JSON compliance rates are recorded.

Configurations are compared by analyzing:
\begin{itemize}
    \item Macro-F1 for overall performance,
    \item per-category F1 for category-level stability,
    \item JSON compliance for output validity.
\end{itemize}

At each successive halving stage, configurations that violate the category-aware acceptance criteria or fail JSON compliance are discarded. Retained configurations are re-evaluated on progressively larger subsets of data, yielding a refined comparison of prompt batch sizes under identical experimental conditions.

\subsubsection{\textbf{Evaluation of Prompt Versions}}

Each prompt version is evaluated as an independent configuration. For a fixed model and batch size, the pipeline is re-run and Stage-wise F1, Macro-F1, per-category F1, and JSON compliance are recomputed. Prompt versions are compared by examining changes in these metrics. A new prompt is retained only if it improves or preserves overall and category-level performance within the acceptance thresholds.

\subsubsection{\textbf{Qualitative Error Analysis}}

For each experiment, misclassified sentences are analyzed qualitatively. Errors are grouped into false positives and false negatives for Stage~1 and category confusions for Stage~2. These analyses identify systematic failure modes and support interpretation of quantitative results.

Overall, the evaluation protocol combines stage-wise metric reporting, controlled comparison of prompt batch sizes, systematic assessment of prompt versions, and structured error analysis. This framework enables reproducible benchmarking of LLMs for clinical action extraction in terms of accuracy, stability, and clinical reliability.

\section{Results and Experiments}

%------------Per-category Precision, Recall, and F1 for CLIP baselines
\begin{table*}[t]
\centering
\caption{Stage 1 (Actionability Filtering): F1 Performance Comparison}
\label{tab:stage1_prf}
\scriptsize
\setlength{\tabcolsep}{3pt}
\renewcommand{\arraystretch}{1.05}
\begin{tabular}{l|c|c|ccc|ccc|ccc|ccc|ccc}
\hline
\textbf{Metric} 
& \multicolumn{2}{c|}{\textbf{BERT Models}}
& \multicolumn{15}{c}{\textbf{LLM Models}} \\
\cline{2-18}
& \textbf{DNote+Cxt} 
& \textbf{TTP+Cxt} 
& \multicolumn{3}{c|}{\textbf{MedGemma-27B-it}} 
& \multicolumn{3}{c|}{\textbf{GPT-5.2}} 
& \multicolumn{3}{c|}{\textbf{Gemini-3-Flash}} 
& \multicolumn{3}{c|}{\textbf{DeepSeek-V3.2}} 
& \multicolumn{3}{c}{\textbf{Claude 3.5 Sonnet}} \\
\cline{2-18}
& \textbf{F1}
& \textbf{F1}
& \textbf{P} & \textbf{R} & \textbf{F1} 
& \textbf{P} & \textbf{R} & \textbf{F1} 
& \textbf{P} & \textbf{R} & \textbf{F1} 
& \textbf{P} & \textbf{R} & \textbf{F1} 
& \textbf{P} & \textbf{R} & \textbf{F1} \\
\hline
\textbf{Binary F1}
& \cellcolor{secondorange}0.856
& \cellcolor{bestgreen}\textbf{0.866}
& 0.379 & 0.582 & 0.459 
& 0.844 & 0.785 & \cellcolor{bestgreen}\textbf{0.813} 
& 0.803 & 0.798 & \cellcolor{secondorange}0.801 
& 0.658 & 0.665 & 0.661 
& 0.731 & 0.739 & 0.735 \\
\hline
\textbf{Macro F1}
& \cellcolor{secondorange}0.661
& \cellcolor{bestgreen}\textbf{0.668}
& 0.642 & 0.748 & 0.684
& 0.909 & 0.883 & \cellcolor{bestgreen}\textbf{0.895}
& 0.889 & 0.887 & \cellcolor{secondorange}0.888
& 0.807 & 0.811 & 0.809
& 0.849 & 0.853 & 0.851 \\
\hline
\end{tabular}
\end{table*}

% Stage 2 - Action based category classification
\begin{table*}[t]
\centering
\caption{Stage 2 (Action Category Classification): F1 Performance Comparison}
\label{tab:stage2_prf}
\scriptsize
\setlength{\tabcolsep}{2.5pt}
\renewcommand{\arraystretch}{1.05}
\begin{tabular}{p{3.8cm}|c|c|ccc|ccc|ccc|ccc|ccc}
\hline
\textbf{Action Category}
& \multicolumn{2}{c|}{\textbf{BERT Models}}
& \multicolumn{15}{c}{\textbf{LLM Models}} \\
\cline{2-18}
& \textbf{DNote+Cxt} 
& \textbf{TTP+Cxt} 
& \multicolumn{3}{c|}{\textbf{MedGemma-27B-it}} 
& \multicolumn{3}{c|}{\textbf{GPT-5.2}} 
& \multicolumn{3}{c|}{\textbf{Gemini-3-Flash}} 
& \multicolumn{3}{c|}{\textbf{DeepSeek-V3.2}} 
& \multicolumn{3}{c}{\textbf{Claude 3.5 Sonnet}} \\
\cline{2-18}
& \textbf{F1}
& \textbf{F1}
& \textbf{P} & \textbf{R} & \textbf{F1}
& \textbf{P} & \textbf{R} & \textbf{F1}
& \textbf{P} & \textbf{R} & \textbf{F1}
& \textbf{P} & \textbf{R} & \textbf{F1}
& \textbf{P} & \textbf{R} & \textbf{F1} \\
\hline
Appointment-related follow-up 
& \cellcolor{secondorange}0.882
& \cellcolor{bestgreen}\textbf{0.887}
& 0.50 & 0.54 & 0.52 
& 0.90 & 0.74 & \cellcolor{bestgreen}\textbf{0.812} 
& 0.82 & 0.72 & \cellcolor{secondorange}0.765 
& 0.45 & 0.50 & 0.474 
& 0.73 & 0.76 & 0.747 \\
\hline
Case-specific instructions for patient 
& \cellcolor{secondorange}0.830
& \cellcolor{bestgreen}\textbf{0.841}
& 0.37 & 0.36 & 0.363 
& 0.78 & 0.77 & \cellcolor{bestgreen}\textbf{0.771} 
& 0.75 & 0.75 & \cellcolor{secondorange}0.750 
& 0.58 & 0.58 & 0.581 
& 0.67 & 0.69 & 0.677 \\
\hline
Imaging-related follow-up 
& \cellcolor{bestgreen}\textbf{0.567}
& \cellcolor{secondorange}0.566
& 0.204 & 0.568 & 0.300 
& 0.54 & 0.51 & 0.525 
& 0.49 & 0.48 & 0.485 
& 0.556 & 0.517 & \cellcolor{bestgreen}\textbf{0.536} 
& 0.51 & 0.48 & \cellcolor{secondorange}0.492 \\
\hline
Lab-related follow-up 
& \cellcolor{secondorange}0.744
& \cellcolor{bestgreen}\textbf{0.745}
& 0.52 & 0.53 & 0.527 
& 0.73 & 0.74 & 0.733 
& 0.885 & 0.662 & \cellcolor{bestgreen}\textbf{0.754} 
& 0.60 & 0.60 & 0.601 
& 0.695 & 0.699 & \cellcolor{secondorange}0.697 \\
\hline
Medication-related follow-up 
& \cellcolor{secondorange}0.659
& \cellcolor{bestgreen}\textbf{0.668}
& 0.377 & 0.326 & 0.350 
& 0.42 & 0.39 & 0.405 
& 0.43 & 0.424 & \cellcolor{secondorange}0.427 
& 0.31 & 0.316 & 0.313 
& 0.48 & 0.669 & \cellcolor{bestgreen}\textbf{0.559} \\
\hline
Procedure-related follow-up 
& \cellcolor{bestgreen}0.597
& \cellcolor{secondorange}\textbf{0.548}
& 0.46 & 0.464 & 0.462 
& 0.50 & 0.48 & 0.491 
& 0.48 & 0.472 & 0.476 
& 0.506 & 0.494 & \cellcolor{bestgreen}\textbf{0.500} 
& 0.49 & 0.477 & \cellcolor{secondorange}0.483 \\
\hline
Other helpful contextual information 
& \cellcolor{secondorange}0.349
& \cellcolor{bestgreen}\textbf{0.365}
& 0.18 & 0.165 & 0.172 
& 0.405 & 0.180 & \cellcolor{bestgreen}\textbf{0.250} 
& 0.21 & 0.16 & \cellcolor{secondorange}0.182 
& 0.12 & 0.116 & 0.118 
& 0.14 & 0.13 & 0.135 \\
\hline
\textbf{Macro Average}
& \cellcolor{secondorange}0.631
& \cellcolor{bestgreen}\textbf{0.668}
& 0.373 & 0.422 & 0.385
& 0.611 & 0.544 & \cellcolor{bestgreen}\textbf{0.570}
& 0.581 & 0.524 & \cellcolor{secondorange}0.548
& 0.446 & 0.446 & 0.446
& 0.531 & 0.558 & 0.541 \\
\hline
\end{tabular}
\end{table*}

We benchmark our proposed two-stage LLM framework against two strong BERT-based supervised baselines from the original CLIP work~\cite{mullenbach-etal-2021-clip}: \textbf{MIMIC-DNote-BERT+Context} and \textbf{TTP-BERT+Context (250k)}. These models represent the state-of-the-art for supervised clinical action extraction on the CLIP dataset and serve as our primary performance reference. For LLM evaluation, we evaluate five models: GPT-5.2~\cite{gpt_5_2}, Gemini-3-Flash~\cite{gemini3_flash}, Claude Sonnet 3.5~\cite{claude35_sonnet}, DeepSeek-V3.2~\cite{deepseekv32}, and the medical-specialized MedGemma-27B-it~\cite{medgemma,sellergren2025medgemmatechnicalreport}. For our experiments, we used $k=15$ as the prompt batch size configuration. The CLIP benchmark test set comprises 100 discharge summaries; our evaluation uses the complete available test set in its entirety, not a subsample. The 518-document figure referenced in Section IV refers to the training split used during prompt batch size selection.  Tables~\ref{tab:stage1_prf} and~\ref{tab:stage2_prf} report precision, recall, and F1 scores for all models across both stages. We highlight the best-performing model in green and second-best in orange within each category (BERT and LLMs). The dataset access was conducted in compliance with the applicable data use agreement and Zero Data Retention (ZDR) requirements, in accordance with PhysioNet’s guidelines for the responsible use of MIMIC-III data.

\subsection{Stage 1: Actionability Filtering}

Table~\ref{tab:stage1_prf} presents actionability filtering results. The BERT baselines achieve strong performance, with TTP-BERT+Context (250k) obtaining a Binary F1 of 0.866 and Macro F1 of 0.668, while MIMIC-DNote-BERT+Context achieves Binary F1 of 0.856 and Macro F1 of 0.661. Among LLMs, GPT-5.2 achieves the strongest performance with a Macro F1 of 0.895, representing a 23\% relative improvement over TTP-BERT+Context (0.668). Gemini-3-Flash achieves the second-best LLM performance with Macro F1 of 0.888. Notably, the medical-specialized MedGemma-27B-it substantially underperforms with  F1 of only 0.459. 

\subsection{Stage 2: Action Category Classification}

Table~\ref{tab:stage2_prf} presents fine-grained classification results across seven clinical action categories. Here, the supervised BERT baselines demonstrate clear advantages, with TTP-BERT+Context (250k) achieving a Macro F1 of 0.668 - the highest overall performance across all models. MIMIC-DNote-BERT+Context achieves Macro F1 of 0.631. Among LLMs, GPT-5.2 achieves the best Macro F1 of 0.570, falling 15\% short of TTP-BERT+Context. GPT-5.2 shows category-specific strengths:
\begin{itemize}
    \item \textit{Appointment-related follow-up}: F1 of 0.812, approaching but not exceeding TTP-BERT's 0.887
    \item \textit{Case-specific instructions}: F1 of 0.771, competitive with TTP-BERT's 0.841
\end{itemize}

Gemini-3-Flash achieves the second-best LLM Macro F1 of 0.548 and excels in \textit{Lab-related follow-up} with F1 of 0.754, matching TTP-BERT's 0.745, while  Claude Sonnet 3.5 demonstrates the best LLM performance for \textit{Medication-related follow-up} with F1 of 0.559. MedGemma-27B-it continues to underperform. DeepSeek-V3.2 shows strengths in:
\begin{itemize}
    \item \textit{Imaging-related follow-up}: F1 of 0.536, approaching TTP-BERT's 0.566
    \item \textit{Procedure-related follow-up}: F1 of 0.5, competitive with MIMIC-DNote-BERT's 0.597
\end{itemize}

% \subsection{Prompt Batch Size Selection}
% \textbf{
% For our experiments, we used 15 as the prompt batch size configuration. }

\section{Ground-Truth Annotations Error Analysis}

Qualitative error analysis reveals that many LLM misclassifications reflect semantic correctness conflicting with dataset annotation conventions rather than true clinical reasoning errors. Three patterns emerged:

\subsection{Semantic vs. Structural Annotation} A notable source of false positives arises from the dataset's reliance on document structure over semantic content. In the original CLIP dataset documentation, the ``\textit{Case-specific patient instructions"} label was standardized by automatically applying it to \textit{all} sentences within ``Followup instructions" and ``Discharge instructions" sections. This strict structural constraint frequently conflicts with the semantic reasoning of LLMs.

For example, models correctly identify imperative sentences such as ``***STOPPED: Lasix, please stop taking this medication" as multi label category actions: they are both \textit{Medication-related follow-ups} (stopping a drug) and \textit{Case-specific patient instructions} (a direct command). However, when such sentences appear within a "Discharge Instructions" block, the ground truth often assigns \textit{only} the \textit{``Case-specific patient instructions"} label, adhering strictly to the section header rule while omitting the semantic medication label. Consequently, the model's inclusion of the semantically accurate \textit{Medication-related} label is penalized as a false positive. This illustrates a key limitation where the dataset's structure-over-semantics approach biases evaluation against models for identifying the multi-dimensional nature of clinical actions.

\subsection{Inconsistency in Medication Follow-up Annotation}
For instance, \textit{doc\_id} 1799 reveals annotation inconsistencies that confound model learning. In the discharge medication list (sentences 106-114), only antibiotics are labeled as \textit{Medication-related followups}:
\begin{itemize}
    \item \textbf{Annotated:} Levofloxacin (line 110), Flagyl (sentence 114)
    \item \textbf{Not annotated:} Aspirin, Enalapril, Prostigmin, Oxycodone, Combivent, Tylenol
\end{itemize}

This pattern initially suggests a rule: acute treatments requiring completion monitoring are actionable, while chronic maintenance medications are not. However, sentence 76 - ``Eventually, ampicillin was discontinued, and the patient was continued on levo and Flagyl" - contradicts this interpretation. This sentence appears in the ``Hospital Course" section, describes past inpatient antibiotic changes using past-tense verbs, and contains no post-discharge instruction, yet is annotated as \textit{Medication-related followups}. Sentence 85 - ``He was continued on an aspirin and an ACE" - uses identical grammatical structure but remains unlabeled.

Sentence 76 represents completed inpatient care, not future follow-up. This suggests either: (1) the annotation guidelines were applied inconsistently, combining medication-type cues with temporal relevance, or (2) the ground truth may contain labeling inaccuracies, causing non-actionable historical information that is correctly identified by LLMs to be evaluated as false negatives.

\subsection{Context Fragmentation and ``Implicit" Actions} Our framework processes documents in fixed batches (k = 15) to manage context window limitations, which can occasionally separate section headers from their associated content. This results in false negatives when LLMs fail to recognize a list of medications as \textit{Medication-related follow-up} because the corresponding header (e.g., “Discharge Medications”) appears in a previous batch.

Furthermore, LLMs struggle with the classification of implicit future actions that are expressed using past-tense language. For example, the sentence “Warfarin anticoagulation was initiated with a goal INR between 1.5–2.5” is frequently missed by the model. Syntactically, “was initiated” describes a completed inpatient action. Clinically, however, it implies a mandatory \textit{future} follow-up task, namely continued ``INR monitoring". The models often filter such sentences out during Stage 1 (Actionability Detection) because they lack explicit future-oriented directives (e.g., “Patient must monitor INR”), highlighting the challenge of identifying actions that are clinically implied but grammatically framed as historical events in the discharge summary.

These patterns suggest that the reported F1 scores may underestimate the true semantic and clinical reasoning capabilities of the models. A hybrid evaluation strategy that combines strict alignment with annotation conventions and assessments of clinical correctness would provide a more faithful measure of model utility for real-world deployment.

\section{Discussion}

\subsection{LLMs vs. Supervised Baselines: A Stage-Dependent Story}

Our results show clear stage-dependent performance patterns. At Stage 1 (actionability filtering), modern LLMs outperform supervised baselines: GPT-5.2 achieves the strongest Stage 1 performance with a Macro F1 of 0.895, a 0.227 point absolute gain over TTP-BERT+Context (0.668). However, at Stage 2 (fine-grained category classification), supervised BERT baselines remain stronger (TTP-BERT+Context Macro F1: 0.668 vs. GPT-5.2: 0.570), suggesting that LLMs are effective at binary filtering but less consistent on multi-label classification that may improve from task-specific training. Notably, LLMs are competitive for some categories---Gemini-3-Flash matches TTP-BERT on \textit{lab-related follow-up} (F1 0.754 vs. 0.745), indicating certain action types are more amenable to zero-shot prompting. Overall, multi-label classification pose the primary challenge across all the models and highlights the need for further investigation into modeling strategies and evaluation design to improve F1 scores.

\subsection{The Medical-Specialized Model Paradox}
MedGemma-27B's consistent underperformance (Stage 1 F1: 0.459, Stage 2 Macro F1: 0.385) relative to general-purpose LLMs reveals a nuanced insight. Although we use MedGemma-27B-it, the instruction-tuned variant, its fine-tuning corpus likely emphasized clinical QA and diagnostic reasoning rather than structured extraction with constrained JSON output schemas. General-purpose models benefit from broader RLHF coverage including tool use and structured output generation, which translates directly to stronger instruction-following in multi-step prompt chaining. MedGemma's low precision with moderate recall (e.g., P = 0.204, R = 0.568 for \textit{Imaging-related follow-up}) reflects this difficulty in distinguishing clinically informative statements from executable instructions. Whether fine-tuning on extraction-specific data would close this gap remains an open question for future work.

\subsection{Clinical Implications}

The precision–recall profiles highlight clinically meaningful trade-offs. Claude Sonnet’s higher medication-related recall (0.669 vs. GPT-5.2’s 0.39) reflects a bias toward capturing more true cases at the cost of increased false positives. In real-world deployment, this reduces the risk of missing medication-related follow-up actions that could impact patient safety. In contrast, Gemini-3-Flash’s high precision (0.885) for lab-related follow-up minimizes spurious orders and helps reduce unnecessary clinical workload. These results suggest that category-specific model selection or threshold tuning may better align model behavior with high risk categories.

However, in comparison to LLMs the higher performance of supervised models in Stage 2 indicates that hybrid architectures—where LLMs handle actionability filtering followed by fine-tuned classifiers for category assignment—may best combine the strengths of both approaches.

\section{Limitations and Future Work}

\textbf{Limitations:} This study evaluates large language models using only zero-shot and few-shot prompting. As a result, it does not assess the impact of fine-tuning approaches, such as parameter-efficient fine-tuning, which may improve performance on task-specific classification. In addition, our error analysis indicates that some observed performance limitations may be driven by inconsistencies in the ground-truth annotations, particularly for cases involving implicit medication logic and structural labeling rules. Consequently, the reported metrics may reflect annotation noise in addition to true model limitations.

Our evaluation is conducted on the CLIP dataset, which draws exclusively from MIMIC-III critical care notes. These results may not generalize to other institutions with different documentation styles or clinical workflows. In addition, the consistently low performance on the \textit{Other helpful contextual information} category suggests that this label is too broadly defined for reliable extraction, reducing its usefulness for quantitative evaluation and indicating a need for taxonomy refinement. We also note that results are reported from single experimental runs. As all LLM inference was conducted at temperature = 0, outputs are deterministic and repeated runs would yield identical results; variance across runs is therefore zero by construction. For non-deterministic settings, future work should report mean and standard deviation across multiple runs to ensure statistical robustness.

\textbf{Future work:} 1) improve annotation consistency to better isolate model behavior from labeling artifacts. Incorporating human-in-the-loop (HITL) validation, including agreement measures such as Cohen’s Kappa between model predictions and clinician judgments, would help distinguish genuine model errors from annotation ambiguities. 2) explore fine-tuning strategies and assess whether smaller language models (SLMs), with or without fine-tuning, can achieve comparable performance to large language models for clinical action extraction. 3) Expanding evaluation to additional datasets beyond CLIP would further support testing robustness across diverse healthcare settings. 4) Finally, evaluating alignment between model reasoning and clinician decision-making processes by comparing model-generated rationales with expert clinical reasoning, rather than relying solely on label-level agreement as it may provide deeper insight into the reliability of model predictions in patient safety–critical applications.

\section {Conclusion}
Our work demonstrates that modern LLMs exhibit strong semantic and clinical reasoning capabilities for actionability detection, achieving performance comparable to and exceeding supervised baselines at Stage 1, while fine-grained category classification at Stage 2 remains an area where task-specific supervised models retain a meaningful advantage, even under strict privacy constraints and without task-specific fine tuning. Qualitative analysis indicates that a substantial portion of observed errors arises from mismatches between model reasoning and dataset annotation conventions, particularly in handling implicit actions and structurally driven labeling rules. These findings suggest that measured performance reflects not only model limitations but also the quality and clarity of the underlying annotation process. The path forward requires parallel advances on three fronts. First, future annotation efforts should capture not only labeled spans but also the clinical rationale justifying each decision, enabling precise diagnosis of model errors. Second, evaluation protocols must distinguish between clinically consequential errors (missed critical follow-ups) and benign interpretation differences (debatable edge cases). Third, deployment architectures should leverage LLM flexibility for initial extraction while routing uncertain cases to specialized classifiers or human reviewers, creating a reliability gradient appropriate for safety-critical applications.

These directions extend beyond discharge note extraction. Many clinical NLP tasks like diagnosis extraction, treatment recommendation identification, and adverse event detection face similar evaluation challenges as models develop sophisticated reasoning capabilities that may not align with surface-level annotation patterns. By acknowledging this measurement problem and developing reasoning-transparent evaluation frameworks, we can build clinical AI systems that are not only accurate by benchmark standards but trustworthy in practice.

\section{Acknowledgement}
The authors thank Beth Israel Deaconess Medical Center for their continued support of the MIMIC project. The authors also thank the MIT Laboratory for Computational Physiology and PhysioNet for providing access to the MIMIC databases. This work used data from the MIMIC-III Clinical Database~\cite{PhysioNet-mimiciii-1.4_database, PhysioNet-mimiciii-1.4_paper}.

\bibliographystyle{IEEEtran}
\bibliography{references}

\end{document}